\documentclass[acmsmall,authorversion]{acmart}
\usepackage{multirow}
\usepackage{subcaption}
\usepackage{siunitx}
\usepackage{bm}
\AtBeginDocument{%
  \providecommand\BibTeX{{%
    \normalfont B\kern-0.5em{\scshape i\kern-0.25em b}\kern-0.8em\TeX}}}

\newcolumntype{C}{>{\FormatNo} c }
\def\FormatNo\ignorespaces#1\\{%
  \ignorespaces\makebox[\widthof{000}][r]{#1}\tabularnewline}

\setcopyright{acmlicensed}
\acmJournal{PACMCGIT}
\acmYear{2024} \acmVolume{7} \acmNumber{1} 
\acmArticle{21} 
\acmMonth{5} 
\acmDOI{10.1145/3651289}

\begin{document}
\title{Efficient Visibility Approximation for Game AI using Neural Omnidirectional Distance Fields}

\author{Zhi Ying}
\email{zhi.ying@ubisoft.com}
\orcid{0009-0008-8390-3366}
\affiliation{%
  \institution{Ubisoft La Forge}
  \city{Shanghai}
  \country{China}
}

\author{Nicholas Edwards}
\email{nicholas.edwards@ubisoft.com}
\orcid{0009-0007-4621-541X}
\affiliation{%
  \institution{Ubisoft}
  \city{Montreal}
  \country{Canada}
}

\author{Mikhail Kutuzov}
\email{mikhail.kutuzov@ubisoft.com}
\orcid{0009-0004-4049-2924}
\affiliation{%
  \institution{Ubisoft}
  \city{Montreal}
  \country{Canada}
}

\begin{abstract}
Visibility information is critical in game AI applications, but the computational cost of raycasting-based methods poses a challenge for real-time systems. To address this challenge, we propose a novel method that represents a partitioned game scene as neural Omnidirectional Distance Fields (ODFs), allowing scalable and efficient visibility approximation between positions without raycasting. For each position of interest, we map its omnidirectional distance data from the spherical surface onto a UV plane. We then use multi-resolution grids and bilinearly interpolated features to encode directions. This allows us to use a compact multi-layer perceptron (MLP) to reconstruct the high-frequency directional distance data at these positions, ensuring fast inference speed. We demonstrate the effectiveness of our method through offline experiments and in-game evaluation. For in-game evaluation, we conduct a side-by-side comparison with raycasting-based visibility tests in three different scenes. Using a compact MLP (128 neurons and 2 layers), our method achieves an average cold start speedup of 9.35 times and warm start speedup of 4.8 times across these scenes. In addition, unlike the raycasting-based method, whose evaluation time is affected by the characteristics of the scenes, our method's evaluation time remains constant.
\end{abstract}



\begin{teaserfigure}
  \includegraphics[width=\textwidth]{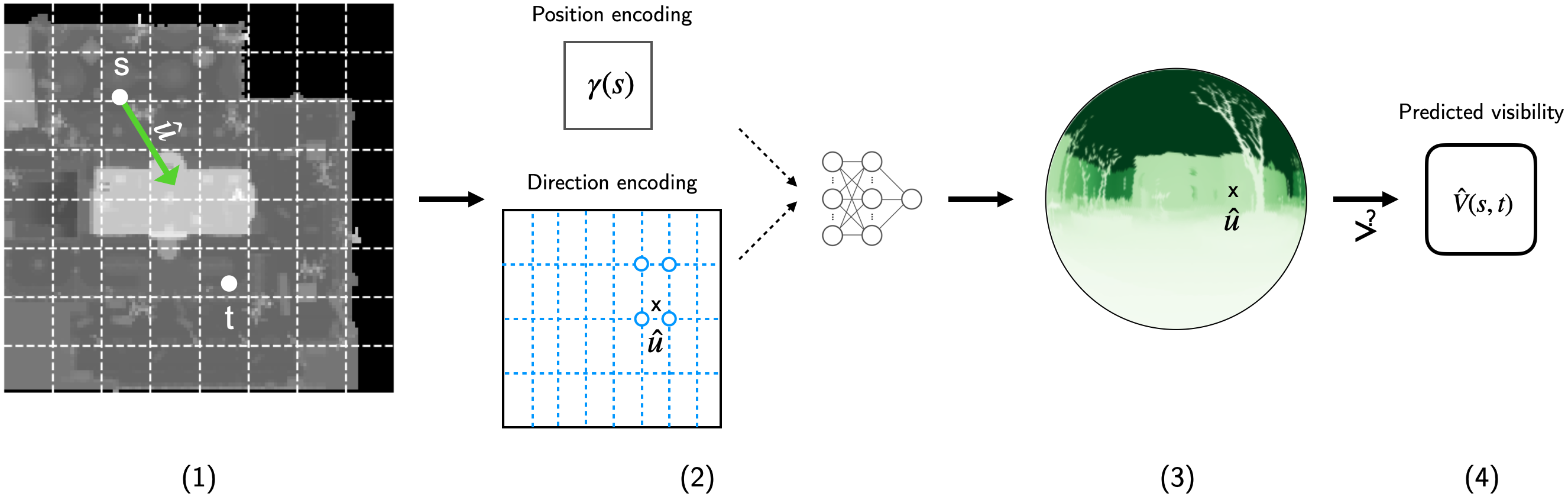}
  \caption{This work presents an efficient approach to approximate visibility between 3D positions in static game scenes using a neural Omnidirectional Distance Field (ODF) representation. It enables game AI entities, such as NPCs, to obtain visibility information without raycasting. The evaluation process of our method is as follows \textbf{(1)} From the bird's-eye view of a 2D grid partitioned game scene, the visibility evaluation takes place between the source position \(s\) and the target position \(t\), where \(\hat{u}\) represents the normalized direction from \(s\) to \(t\). \textbf{(2)} The local ODF representation associated with the partitioned area containing \(s\) is inferred, the positional encoding \(\gamma\) encodes the position, while the multi-resolution grid encodes the direction. These encoded features are concatenated and fed into a MLP. \textbf{(3)} The MLP returns the distance value originating from \(s\) in the direction \(\hat{u}\). \textbf{(4)} The predicted visibility result is computed by comparing the inferred distance with the Euclidean distance between \(s\) and \(t\).}
  \label{fig:teaser}
\end{teaserfigure}


\maketitle

\section{Introduction}
In video games, visibility or line-of-sight information is crucial for AI entities such as non-player characters (NPCs) to build spatial knowledge about the game world \cite{bourg2004ai}. It is often used to make real-time decisions, which can significantly affect the ability of NPCs to appear intelligent. The most basic result of a visibility test is a boolean value describing whether two positions in the game world are visible to each other, i.e. \(V(s, \,t) \in \{0, 1\}\). Advanced systems for building complex and believable behaviors for game AI rely heavily on it \cite{walsh2019modeling, welsh2013crytek, mcintosh2019human, jack2019tactical}.

For game AI, the general method of obtaining visibility information is through raycasting \cite{glassner1989introduction}. Given a ray, this approach determines the first intersection with a scene object. However, the time required increases with the geometric complexity of the scene. Furthermore, since game AI programs typically run in CPU processes, performing a large number of visibility tests can lead to a computational bottleneck.

To compensate for the high computational cost, game developers often limit the number of visibility tests that are processed in real time. However, this trade-off can have a significant impact on the complexity and believability of NPC behavior. Limiting visibility information can lead to simplistic and unrealistic NPC decision making, limiting the range of actions that NPCs can take in response to their environment. It is therefore important to develop more efficient methods for assessing visibility.

Inspired by recent research on modeling 3D objects and scenes with neural implicit representations \cite{mildenhall2020nerf, chibane2020ndf, mueller2022instant, liu2023raydf}, we propose to represent the surface of the game scene geometry with a continuous distance function that can be optimized by parametric encoding and neural networks to effectively approximate visibility. Due to the varying size of game scenes, we employ a partitioning strategy to divide the entire game scene into smaller areas. For each area, we represent it as a local ODF. The ODF can be expressed as a scalar function that returns the distance to the closest surface based on a given source position \(p\) and direction \(\hat{d}\). Formally, it is defined as
\begin{equation}
\label{def_odf}
ODF(p, \,\hat{d}) \coloneqq \min(\lVert p - p^I \rVert_2), \, p^I \in \partial \Omega,
\end{equation}
where \(p^I = p +  \lambda \hat{d}, \,\lambda \geq 0 \). Using this representation, we can efficiently evaluate whether there are occluders between two positions \(s\) and \(t\) and compute their visibility:
\begin{equation}
V(s, \, t) = 
\begin{cases} 
1, & \text{if } ODF_{i}(s, \, \hat{u}) > \lVert s - t \rVert_2 \\
0, & \text{otherwise}
\end{cases}
\quad \text{where} \quad 
\hat{u} = \frac{t - s}{\lVert t - s \rVert_2}
\quad \text{and} \quad
1 \leq i \leq N
\end{equation}

\noindent Here \(N\) is the total number of partitions, \(i\) is the partition index associated with the source position \(s\). The visibility evaluation process depends on only one of these local ODFs, regardless of the location of the target position. This allows us to represent each area of the game scene with a smaller neural network and achieve fast evaluation, it also allows developers to selectively enable ODF representations in disconnected areas of the entire game scene. Obviously there's a trade-off between computational cost and overall memory usage, but game developers can find their optimal configurations based on specific requirements. 

Standard neural implicit representations designed for view synthesis or 3D shape reconstruction require only low-frequency signals in the direction dimensions. In contrast, our ODF representation requires the capture of higher-frequency details. To address this challenge, for each source position, we map the omnidirectional distance data on the unit spherical surface onto a UV plane. On this UV plane, we employ bilinearly interpolated multi-resolution grid encoding as our direction features, complemented by positional encoding as our position features. By concatenating these features and feeding them into a compact MLP, we can reconstruct ODFs for the game scene while maintaining fast evaluation speeds.

In summary, our contributions are threefold:
\begin{itemize} 
  \item We provide evidence that the learned neural ODF representation can be used to approximate the visibility between positions in static game scenes. 
  \item Evaluating visibility using the ODF depends only on the local representation associated with the source position, regardless of the target's location. We show that by partitioning the game scene into areas and representing each area as an independent ODF, our method can be scaled to game worlds of arbitrary size while maintaining evaluation speed.
  \item We show that by mapping omnidirectional distance data from the spherical surface onto a UV plane and using multi-resolution grid encoding, we are able to effectively reconstruct the distance data using a compact MLP for visibility approximation.
\end{itemize}

\section{Background and Related Work}
The core of our method is to replace traditional raycasting-based visibility computation with neural representation evaluation for efficiency. In this section, we provide a brief review of traditional visibility for NPC behaviors and neural implicit representations. To our knowledge, our work is the first to integrate these two domains, demonstrating promising results in a real game environment.

\subsection{Visibility for NPC Behaviors}

In NPC behavior design, visibility refers to knowing whether one position in a game scene can be "seen" by an object at another position. In order to design complex and believable behaviors, various parts of the NPC decision making logic rely heavily on visibility information. This includes higher-level systems such as NPC visual perception \cite{walsh2019modeling, welsh2013crytek, mcintosh2019human} and tactical position selection \cite{jack2019tactical}, as well as more moment-to-moment logic, such as checking whether an agent should pull the trigger, change stance, or take some other immediate actions \cite{mcintosh2019human}. For example, when evaluating NPC movement destinations during combat, visibility information between candidate positions and the player's current position is critical.

While z-buffer scanning and raycasting \cite{bittner2003visibility} are common algorithms for visibility tests in graphics, NPC visibility queries require testing pairs of positions scattered throughout the game scene. Therefore, raycasting-based methods are often more efficient. However, as game scenes become more complex, raycasting algorithms face performance challenges. \cite{smits2005efficiency} highlights the complexity and considerations required to optimize raycasting. Although more efficient raycasting algorithms \cite{moller2005fast}, acceleration structures \cite{klimaszewski1994faster, klosowski1998efficient, foley2005kd}, and hardware-accelerated libraries \cite{ingo2014embree} have been developed, the cost of intensive visibility computation is still significant in many modern video games.

The probe-based irradiance field with visibility representation, as introduced in \cite{mcguire2017realtime, Majercik2019Irradiance}, is a promising advancement in real-time global illumination. By storing and constantly updating visibility information in GPU-resident probes, this technique effectively mitigates rendering problems such as light leaks. Accessing low-detail dynamic visibility results from these probes and estimating them by interpolation can be efficient \cite{Majercik2019Irradiance}. However, the transition to game AI applications presents notable challenges. First, the probes are designed for rendering tasks, so they are often placed around the camera or cover only part of the scene. For game AI applications, however, the probes need to be placed and active for a larger area, which can significantly increase its GPU memory usage. Second, game AI applications operate primarily on the CPU, requiring low throughput visibility tests with minimal latency response. This makes probe-based methods less optimal.

In our approach, we leverage neural representations to approximate visibility and execute in the CPU. This is because achieving 100\% accuracy for each visibility test is often unnecessary in game AI applications. For example, the visibility from a position to an object is often evaluated by sampling multiple visibility tests against the volume of that object, so the sampling errors can be handled by statistical techniques.

\subsection{Neural Implicit Representations}

Neural implicit representations have emerged as a powerful tool for encoding high-dimensional data as a continuous, implicit function defined by a neural network, enabling various applications such as novel view synthesis (NVS), 3D shape reconstruction, and physical simulation. For example, NVS seeks to render a scene from unseen viewpoints using a dataset of images and camera poses. In particular, Neural Radiance Fields (NeRF) \cite{mildenhall2020nerf} learns a representation of the scene using a continuous volumetric function of color density and relies on ray marching to render view-dependent images. In our task, we are more interested in distance fields that can be used to evaluate visibility.

Previous studies have explored the use of neural distance functions primarily for 3D shape reconstruction, such as predicting unsigned distance fields from sparse point clouds \cite{chibane2020ndf} or learning consistent distance and density fields for improved localization \cite{ueda2022neural}. However, these distance functions only model distance with respect to the surface normal and rely on algorithms such as sphere tracing to reconstruct shape, making them unsuitable for fast visibility evaluation between positions. 

Classical implicit neural representations suffer from slow training and inference due to the computational complexity. To improve efficiency, previous research falls broadly into two categories: accelerating volumetric representations and non-volumetric representations. 

\textit{Accelerating volumetric representations}. With the introduction of auxiliary acceleration data structures, NeRFs can efficiently evaluate samples with a shallow MLP or no MLP at all. Plenoctrees \cite{yu2021plenoctrees}, plenoxels \cite{yu_and_fridovichkeil2021plenoxels} and multiresolution hash encoding \cite{mueller2022instant} use voxel grids or hash table data structures as position features, along with low degree truncated spherical harmonics (SH) as direction features. A typical setting is SH truncated at degree 2 (\(SH_2\)). Sparse Neural Radiance Grid (SNeRG) \cite{hedman2021snerg} bakes NeRF's continuous neural volumetric scene representation into a discrete data structure at training time, and uses a small MLP with view direction at evaluation time for efficient rendering. These methods still rely on ray marching for evaluation and thus require multiple MLP inferences for a single sample. Another notable effort to accelerate volumetric representations is through point-based radiance fields. 3D Gaussian Splatting \cite{kerbl3Dgaussians} represents the scene using point-based 3D Gaussians and optimizes them using differentiable rendering. This allows the representation evaluation to skip unnecessary computations in empty space, resulting in real-time rendering speed.

\textit{Non-volumetric representations}. Various works represent the scene using other representations to be more efficient. MobileNeRF \cite{chen2022mobilenerf}, represents the scene as a triangle mesh textured by learned deep features. This method leverages a classical rasterization pipeline, incorporating a compact MLP implemented as a GLSL fragment shader. The rendering process for each sample involves a single MLP inference. As a result, they achieve interactive frame rates on mobile phones. It shows that the volumetric representations can be baked onto a proxy surface geometry for efficient rendering. The ray-based implicit 3D shape representations \cite{feng2022prif, liu2023raydf} are another type of non-volumetric representation. They have a similar concept to our work, modeling distance fields based on rays. However, their ray parameterizations are tailored for 3D shape reconstruction, which limits their application to single objects or small scenes. In contrast, our focus is on rays that originate from selected sparse source positions and can be easily scaled to cover the entire game world. Our approach models the omnidirectional distance originating from these positions, allowing us to efficiently test occluders between them and arbitrary target positions. In particular, previous work has introduced the concept of neural ODF for 3D reconstruction using recursive inference \cite{houchens2022neuralodf}, the computation time of which is impractical for our task.

Although some NeRFs can handle scenes the size of a single room or building, they are generally still limited and do not scale to large environments such as the game world. Researchers in the field have proposed to extend NeRF using architectures that align multiple NeRFs in a distributed manner, as shown in \cite{tancik2022blocknerf, turki2022mega}. Our approach follows a similar concept of partitioning the entire scene. However, unlike previous methods that use an end-to-end architecture to dynamically select NeRFs for rendering tasks, we exploit the fact that the evaluation of visibility depends only on the local ODF representation. This insight allows us to partition the game scene through a simple grouping of source positions, paving the way for a scalable and adaptable application.

\section{Method}
\label{sec:method}
Our method can be seen as a precomputation approach that learns the scene geometry in an offline process and then evaluates the learned representation online. In this section, we outline the collection of training data, introduce the multi-resolution grid encoding used to reconstruct high-frequency signals in direction dimensions, and describe the details of model training and evaluation.

\begin{figure}
\centering

\begin{subfigure}{0.2\textwidth}
    \includegraphics[width=\textwidth]{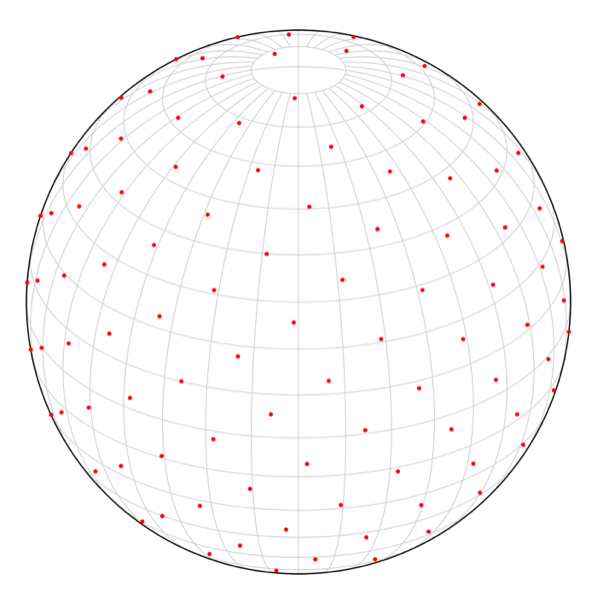}
    \caption{}
\end{subfigure}
\hspace{0.05\textwidth}
\begin{subfigure}{0.4\textwidth}
    \includegraphics[width=\textwidth]{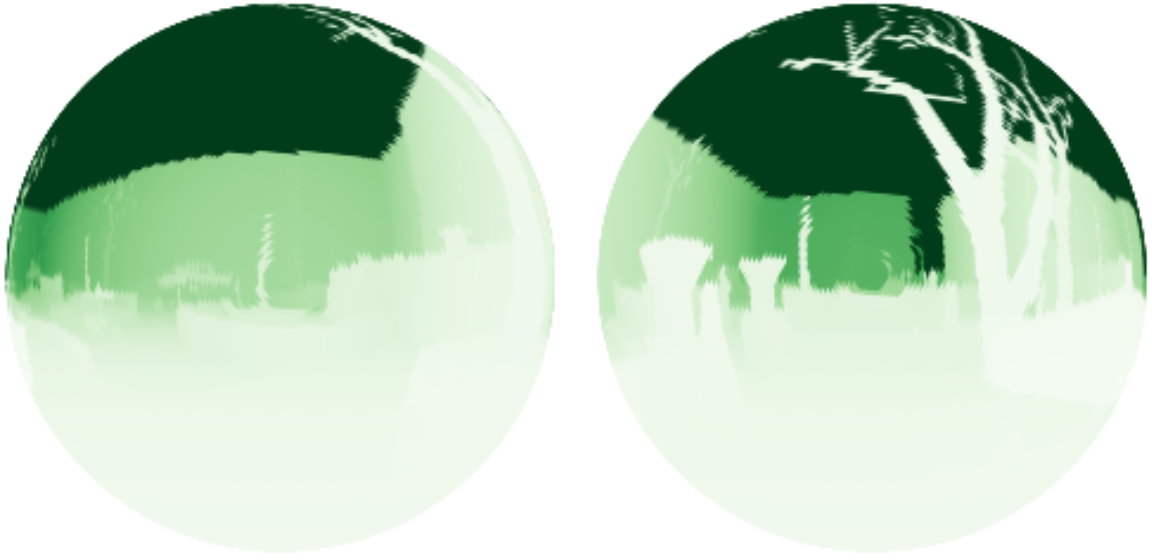}
    \caption{}
\end{subfigure}

\caption{Omnidirectional distance data collection at a source position. \textbf{(a)} Directions are distributed on the spherical surface by a Fibonacci lattice, this visualization shows the total number of directions \(P = 201\). \textbf{(b)} Collected omnidirectional distance data at one position of our offline experiment scene, with \(P = 40001\).}
\label{fig:data_collection}
\end{figure}

\subsection{Training Data Collection}
\label{sec:method:training}
For positions of interest in the game scene, the input variables of ODF can be expressed as rays \(r = (p_r, \,d_r)\), where \(p_r \in \mathbb{R}^3\) represents the position, and \(d_r \in \mathbb{S}^2\) represents the normalized direction. The ODF distance value associated with each ray can be collected in the game scene through raycasting.

In practice, we can only generate a finite number of rays for data collection. To distribute the rays evenly across the surface of the unit sphere,  we employ Fibonacci lattice \cite{swinbank2006fibonacci, gonzalez2010measurement} as a basic approximation. Let \(N\) be any natural number, integer \(i \in [-N,\,+N]\). The \(i\)th direction in radians are
\begin{equation}
\begin{split}
    lat_i &= \arcsin\left( \frac{2i}{2N+1} \right), \\
    lon_i &= 2\pi i\phi^{-1},
\end{split}
\end{equation}
where
\begin{equation}
\begin{gathered}
    \phi = \frac{1+\sqrt{5}}{2} = \lim_{n\to\infty}\left( \frac{F_{n+1}}{F_n} \right), \\
    F_0 = 0, \quad F_1 = 1, \quad
    F_n = F_{n-1} + F_{n-2} \quad \text{for } n > 1
\end{gathered}
\end{equation}
The number of directions is
\begin{equation}
    P = 2N + 1.
\end{equation}

After collecting data at all positions of interest, partitioning can be implemented as a post-processing step. This involves virtually grouping rays based on position \(p_r\), for example, grouping them by the voxels to which \(p_r\) belongs.

\subsection{Multi-resolution Grid Encoding}
\label{sec:method:multi-resolution-grid}

\begin{figure*}[!t]
\centering
\includegraphics[width=\linewidth]{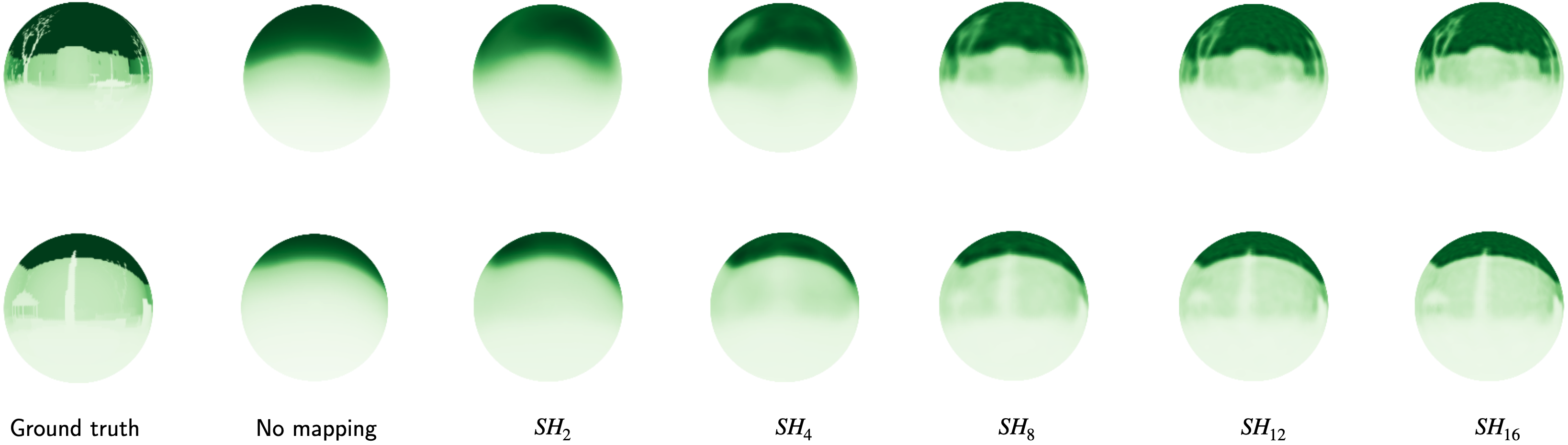}
\caption{ The effect of the high-frequency expressiveness of the neural network due to direction encoding in ODF reconstruction. As the truncated SH degree for direction encoding increases, the reconstructed distance data from the trained neural network contains more details. }
\label{fig:sh_degrees}
\end{figure*}

Standard MLPs are known to have a learning bias toward smooth functions, making them poorly suited for low-dimensional coordinate-based vision and graphics tasks \cite{rahaman2019spectral}. To overcome this limitation, neural representations employ encoding methods that map inputs to higher dimensions, improving the expressiveness for high-frequency signals. Unlike other representations designed for volumetric rendering or 3D shape reconstruction \cite{yu2021plenoctrees, yu_and_fridovichkeil2021plenoxels, mueller2022instant}, whose common practice is to use lower degree truncated SH to encode directions, e.g. \(SH_2\), our neural ODF representation requires capturing high frequency details for directions. This is demonstrated by an experiment of ODF reconstruction at a single position. In the experiment, we set several truncated SH degrees to encode directions, the greater the truncated SH degree, the more SH coefficients are used, the more high-frequency details the trained neural network can capture, as shown in Fig. \ref{fig:sh_degrees}.

However, high-degree SH encoding is computationally expensive. The direct computation scales \(\mathcal{O}(N^3)\) \cite{muller2006spherical}. Although there are some methods to reduce the computational cost \cite{suda2002fast, dutordoir2020sparse}, we decided to develop a more efficient approach.

\begin{figure}
\centering

\begin{subfigure}{0.525\textwidth}
    \includegraphics[width=\textwidth]{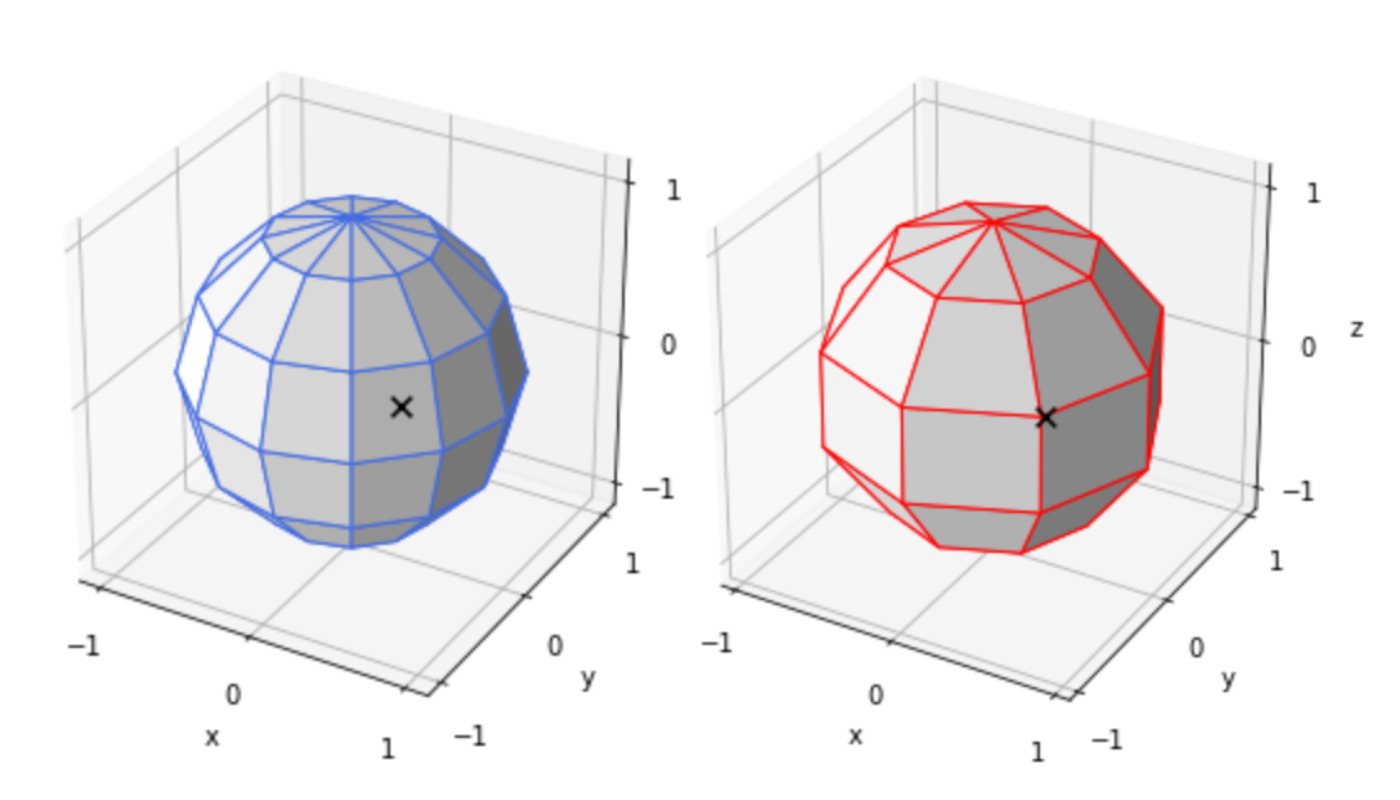}
    \caption{Spherical view}
\end{subfigure}
\begin{subfigure}{0.3\textwidth}
    \includegraphics[width=\textwidth]{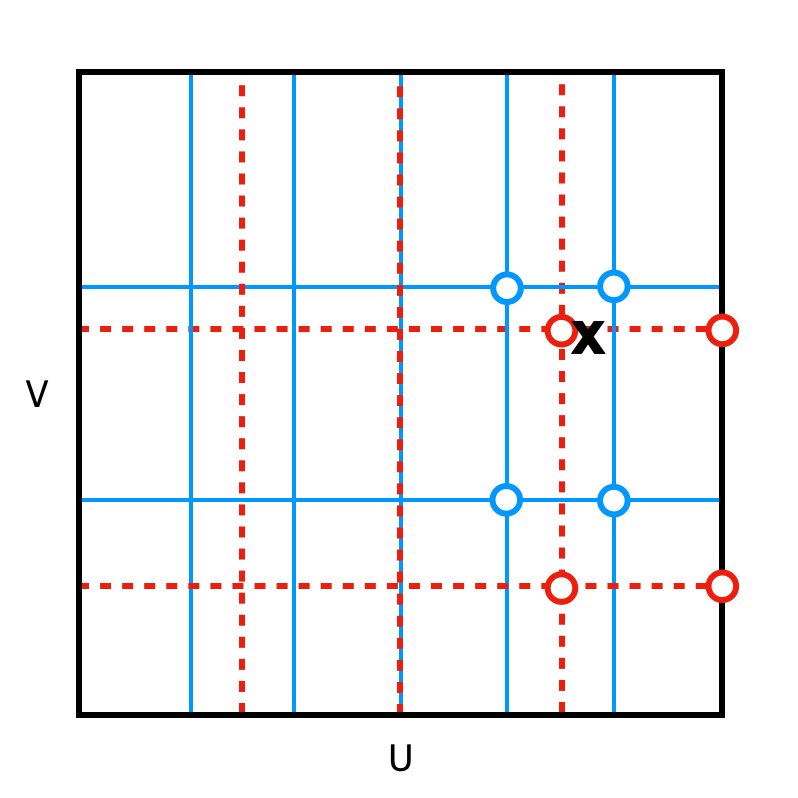}
    \caption{UV plane view}
\end{subfigure}

\caption{Illustration of multi-resolution grid encoding. \textbf{(a)} Spherical view of a two-level resolution grid in the longitude-latitude projection, where the sample direction is represented as "\(\mathbf{x}\)". \textbf{(b)} The (partial) UV plane view with the same grid and direction. The four nearest texels that store relevant features are highlighted for each grid.}
\label{fig:mult-resolution_grid}
\end{figure}

To address the encoding challenge while aiming for fast evaluation, we consider the onmidirectional distance data as a 2D spherical surface, so that it can be mapped onto the UV plane for 2D reconstruction. Inspired by the trade-off between memory and evaluation time made by discretized volumetric representations \cite{hedman2021snerg, mueller2022instant}, we employ a multi-resolution grid data structure to store our parametric direction features. Each grid is rectangular in shape and has a ratio of approximately \(2:1\) for the longitude and latitude axes to match their value ranges. The grid settings we used in our offline experiments are the coarsest resolution of \(16 \times 8\) and the finest resolution of \(256 \times 128\), with the number of levels \(L=16\) and the length of the feature stored in each texel \(F=5\). For each ray direction \(d_r\), the direction features are bilinearly interpolated from the four nearest texels of the UV plane and concatenated across multiple resolutions of the grid. We illustrate a two-level resolution grid example in Fig. \ref{fig:mult-resolution_grid}.

\subsection{Learning and Evaluating ODF}
\label{sec:method:learning_evaluating}
In addition to the multi-resolution grid used for direction encoding, we map the position \(p_r\) to higher dimensional space using hash encoding \cite{mueller2022instant} or positional encoding (PE) \cite{mildenhall2020nerf}, these features are then concatenated and finally fed into a compact MLP. To optimize the neural network, we perform direct regression of the ground truth distance, minimizing the mean squared error. During the optimization, we use the Adam optimizer. To prevent divergence, we apply weak weight decay regularization (coefficient \(10^{-5}\)) to the MLP.

Directly evaluating performance on randomly split data is not ideal due to ray aliasing \cite{feng2022prif}. If we consider moving the origin position \(p_r\) of a ray along it's direction \(\hat{d_r}\), the new ray \(r' = (p_{r'}, \hat{d_r})\) is an aliased version of \(r\). This aliasing effect can also be observed from the ODF perspective, as highlighted by \cite{houchens2022neuralodf} with the equation:
\begin{equation}
\label{odf_constraint}
\nabla_{[\hat{d}, \,\hat{0}]} ODF(p, \,\hat{d}) = -1,
\end{equation}
we can rewrite it as
\begin{equation}
\label{odf_relation}
ODF(p + \lambda \hat{d}, \,\hat{d})  = ODF(p, \,\hat{d}) - \lambda, \,\lambda \in [- ODF(p, \,-\hat{d}), \,ODF(p, \, \hat{d})].
\end{equation}
\noindent A direct evaluation based on a randomly partitioned training and testing set of these rays could result in overestimated metrics.

To mitigate ray aliasing in performance evaluation, we construct our testing dataset in the game scene separately by simulating visibility tests between candidate NPC movement destination positions and player positions. Each candidate position corresponds to one of the source positions where we collected omnidirectional distance data, while each target position is chosen randomly from the entire game scene. Additionally, we collect the results of raycasting-based visibility tests as the ground truth labels, denoted as \(y\). The testing dataset is represented as:
\begin{equation}
\label{eqn:bench_data}
\mathcal{D} = \{(x^{(i)}, \,y^{(i)})\}_{i=1}^M, \, y \in \{0, 1\},
\, x = (s, \,t), \, s \in \mathcal{P}, \,t \in \mathcal{Q},
\end{equation}
where \(\mathcal{P}\) represents positions of interest and \(\mathcal{Q}\) represents all positions within the game scene.

\section{Experiments}
\label{sec:exp}

\subsection{Offline Experiment}
\label{sec:exp_offline}
\begin{figure}
\centering

\begin{subfigure}{0.35\textwidth}
    \includegraphics[width=\textwidth]{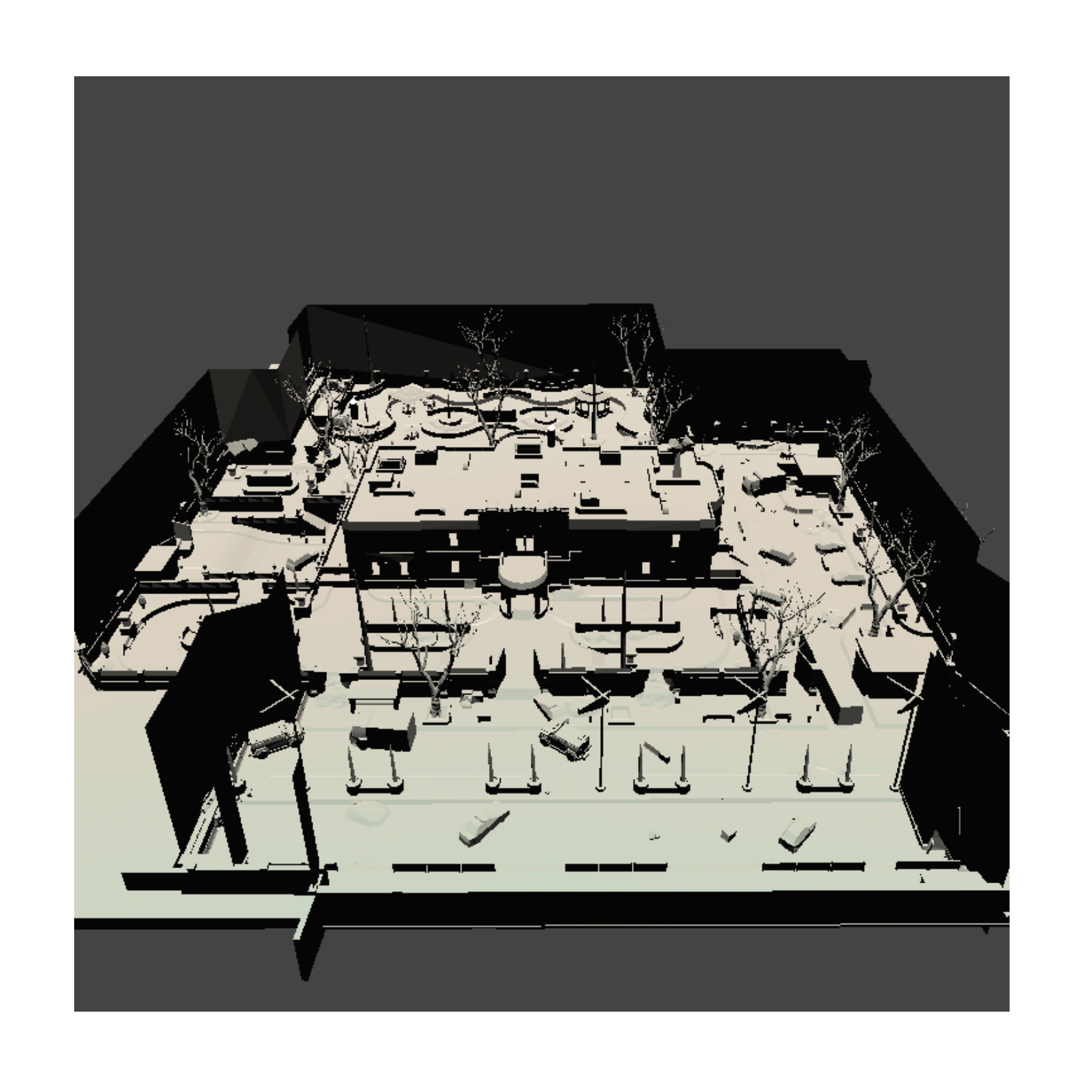}
    \caption{Perspective view}
\end{subfigure}
\hspace{0.05\textwidth}
\begin{subfigure}{0.35\textwidth}
    \includegraphics[width=\textwidth]{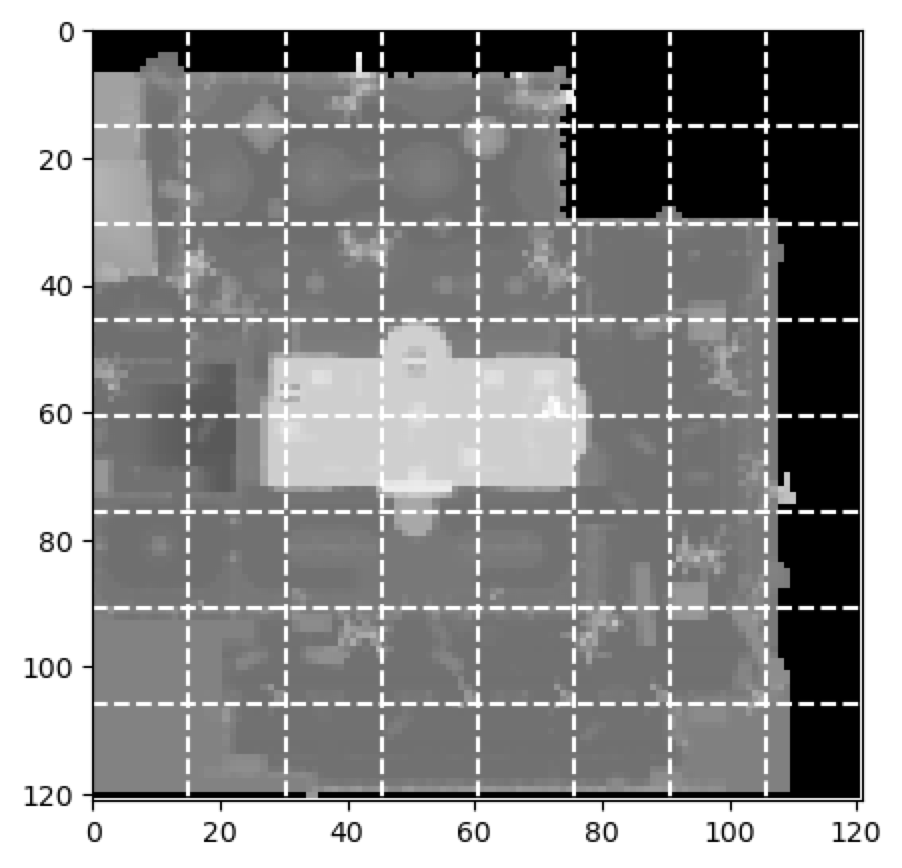}
    \caption{Bird's-eye view with partitioning}
    \label{fig:offline_environment:bev}
\end{subfigure}

\caption{Offline evaluation environment.}
\label{fig:offline_environment}
\end{figure}

To establish an offline environment for convenient performance comparison with baselines, we first collect training and testing data in a game scene using the methods described in Section \ref{sec:method:training} and Section \ref{sec:method:learning_evaluating}. The data is exported as binary files for convenient integration with machine learning frameworks such as PyTorch. We then post-process the training data by partitioning the 6227 source positions using a simple 2D grid scheme, as shown in Fig. \ref{fig:offline_environment:bev}. Based on this \(8 \times 8\) partitioning scheme, we create 64 independently initialized small models for each method under evaluation.  
During both the training and testing phases, each sample infers only one of these models, determined by the local partition index associated with the given source position.

\begin{table}
\footnotesize
  \caption{Offline evaluation for encoding methods.}
  \label{tab:pred_perf}
  \begin{tabular}{l|llll|rr}
    \toprule
     {} & $\uparrow$ Acc. & $\uparrow$ Prec. & $\uparrow$ Recall & $\uparrow$ F1 & $\downarrow$ Params \# (k) & $\downarrow$ Time (ms) \\
    \midrule
    No mapping          & 0.8721 & 0.9032 & 0.7151 & 0.7982 & \textbf{50}  & \bm{$0.27 \pm 0.03$} \\
    Positional encoding & 0.8913 & 0.9136 & 0.7651 & 0.8328 & 54  & $0.41 \pm 0.02$ \\
    Hash encoding ($SH_{2}$)& 0.8750 & 0.9054 & 0.7222 & 0.8035 & 2152 & $6.44 \pm 0.19$ \\
    Hash encoding ($SH_{12}$)& 0.9029 & 0.9303 & 0.7843 & 0.8511 & 2172 & $33.54 \pm 4.94$ \\
    FFM ($SH_{12}$)& 0.8975 & 0.9272 & 0.7708 & 0.8418 & 104 & $29.70 \pm 0.47$ \\
    Ours (Hash, Long-Lat) & 0.9079   & 0.9328 & 0.7971 & \textbf{0.8596} & 3118 & $12.66 \pm 0.55$ \\
    Ours (Hash, Mercator)     & \textbf{0.9081}        & \textbf{0.9336} & 0.7940 & 0.8594 & 3118 & $12.88 \pm 0.37$ \\
    Ours (PE, Long-Lat) & 0.9034   & 0.9167 & \textbf{0.7996} & 0.8542 & 1022 & $6.74 \pm 0.11$ \\
    Ours (PE, Mercator)     & 0.9031            & 0.9221 & 0.7931 & 0.8528 & 1022 & $7.00 \pm 0.16$ \\
    \bottomrule
  \end{tabular}
\end{table}

The evaluation results are presented in Table \ref{tab:pred_perf}. All these experiments are performed offline with PyTorch. The prediction accuracy is evaluated through classification metrics \cite{powers2008evaluation}. The inference time is benchmarked on a Xeon W-2135 CPU and is constrained to single-threaded execution without batching, simulating the typical execution of visibility tests in game AI applications.

In the positional encoding \cite{mildenhall2020nerf} experiment, we use 6 exponential growth frequencies (\(n_{\text{freq}} = 6\)) for both position and direction inputs. In the hash encoding \cite{mueller2022instant} experiment, we use the setting \((F=2, \,L=16)\) with \(2^{16}\) as the hash table size. For its direction encoding we employ SH truncated at degree 2 and degree 12. In the Fourier Feature Mapping (FFM) \cite{tancik2020fourfeat} experiment, we use a standard deviation \(\sigma = 1.0 \) for the Gaussian distribution to encode positions and SH truncated at degree 12 to encode directions. In our method, for directions, we experimented with longitude-latitude projection and the Mercator projection for spherical mapping, the multi-resolution grid setting is described in Section \ref{sec:method:multi-resolution-grid}, which is \((F=5, \,L=16)\). For positions, we use hash encoding and PE of \(n_{\text{freq}}=6\). The hidden size of the MLPs is 128 neurons and 4 layers for all methods.

For all methods, there is a trade-off between prediction performance, memory, and computational cost. From the result, our method with PE shows promising metrics in both prediction performance and computational cost, at a reasonable memory cost. With hash encoding, our method achieves slightly better prediction performance but slower speed. We don't see much difference between the two projection methods with the experimented grid settings. Due to the smoothness in reconstructing omnidirectional distance data, hash encoding with SH truncated at degree 2 has a similar prediction performance to the no mapping baseline. In our experiments, we observed that while positional encoding can yield commendable results with appropriate frequency configurations, its effectiveness is highly sensitive, requiring hyperparameter tuning across different scenes. Hash encoding and FFM with SH truncated at degree 12 are much slower due to their SH computations. Among different versions of our methods, we choose the PE with longitude-latitude projection for further in-game experiments due to its simplicity, faster speed, and lower memory cost.

\begin{figure}
\centering
\begin{subfigure}{0.47\textwidth}
    \includegraphics[width=\textwidth]{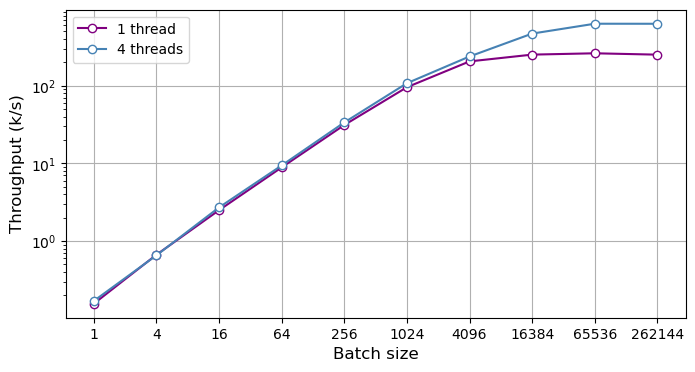}
    \caption{CPU}
\end{subfigure}
\hfill
\begin{subfigure}{0.47\textwidth}
    \includegraphics[width=\textwidth]{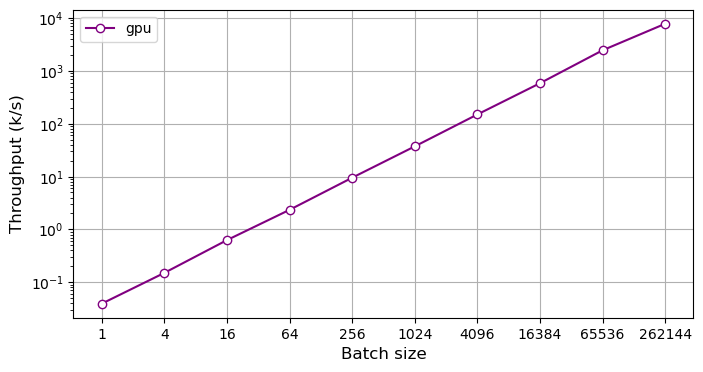}
    \caption{GPU}
\end{subfigure}
        
\caption{Visibility test throughput (kilo tests per second) of our method (PE, Long-Lat) for different batch sizes, tested on CPU and GPU using PyTorch.}
\label{fig:batching_throughput}
\end{figure}

Certain applications require the use of multi-threaded CPU or GPU batching operations to perform high-throughput visibility testing. Our approach benefits from the integration of parallel and vectorized computation with batching. The scalability of the inference throughput of our method for both CPU and GPU is illustrated in Fig. \ref{fig:batching_throughput}. The throughput depends on the hardware and inference framework used; we present the measurements tested using PyTorch, a Xeon W-2135 CPU, and a RTX A5000 GPU.

\subsection{Memory Usage Analysis}

\begin{figure}
\centering
\begin{subfigure}{0.4\textwidth}
    \centering
    \includegraphics[width=\textwidth]{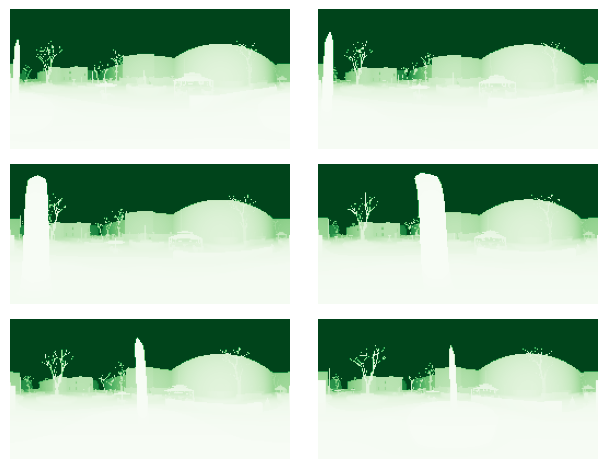}
    \caption{}
\end{subfigure}
\hspace{0.1\textwidth}
\begin{subfigure}{0.395\textwidth}
    \centering
    \includegraphics[width=\textwidth]{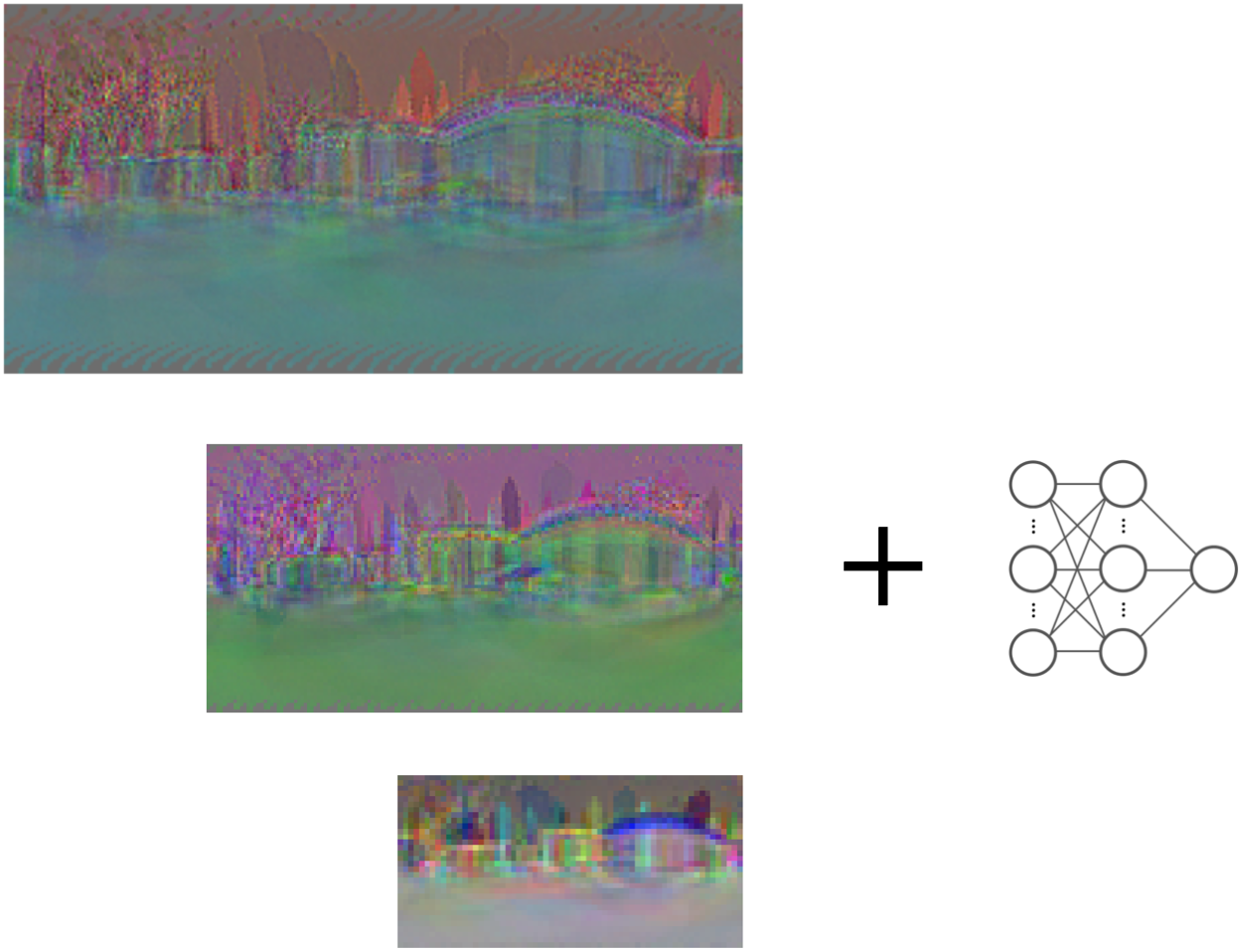}
    \caption{}
\end{subfigure}
        
\caption{Comparison of memory composition between omnidirectional depth maps and our neural ODF representation. \textbf{(a)} Shows omnidirectional depth maps from 6 source positions within a local partition. \textbf{(b)} Shows the neural ODF representation with a 3-level resolution grid for the same partition. The visualization shows the first 3 dimensions of the feature vectors of the grids.}
\label{fig:memory_depth_map_compare}
\end{figure}

Our method involves compressing omnidirectional depth maps for all source positions within the local partition into a single neural ODF representation. Fig. \ref{fig:memory_depth_map_compare} illustrates a comparison of the memory composition. To analyze the difference in memory usage, we conduct an experiment using the offline environment with the partitioning and settings of our method described in Section \ref{sec:exp_offline}. Each partition in the experiment contained an average of 100 source positions. For PNG DEFLATE lossless compression, we used ZLIB compression level 6.

\begin{table}
\footnotesize
  \caption{Memory usage per partition.}
  \label{tab:memory_comparison}
  \begin{tabular}{ll|ll|l}
    \toprule
      \multicolumn{2}{l|}{\textbf{Depth Map} $(256 \times 128)$ }&
      \multicolumn{2}{l|}{\textbf{Depth Map} $(512 \times 256)$ }&
      \multirow{2}{*}{\textbf{Ours}} \\
      {Uncompressed} & {PNG Compressed} & {Uncompressed} & {PNG Compressed} \\
    \midrule
    13.1 MB & 559 KB & 52.4 MB & 1.3 MB & 4.95 MB \\
    \bottomrule
  \end{tabular}
\end{table}

As shown in Table.\ref{tab:memory_comparison}, for each partition, our neural representation has a significantly reduced memory size compared to the size of uncompressed depth maps, although it remains larger than PNG compressed memory size. Traditional image compression methods have their own drawbacks when applied to game AI applications, such as additional computation time, inconvenience in evaluating individual pixels, and lack of generalization ability. To further optimize the memory size of our neural representation, future considerations may include actions such as reducing weight precision to 16-bit float, applying quantization, or using smaller feature vectors. For the scope of this study, our primary focus is on prediction accuracy and inference time in a real game environment, which will be discussed in the following section.

\subsection{In-game Evaluation}

\begin{figure}
\centering
\begin{subfigure}{0.32\textwidth}
    \includegraphics[width=\textwidth]{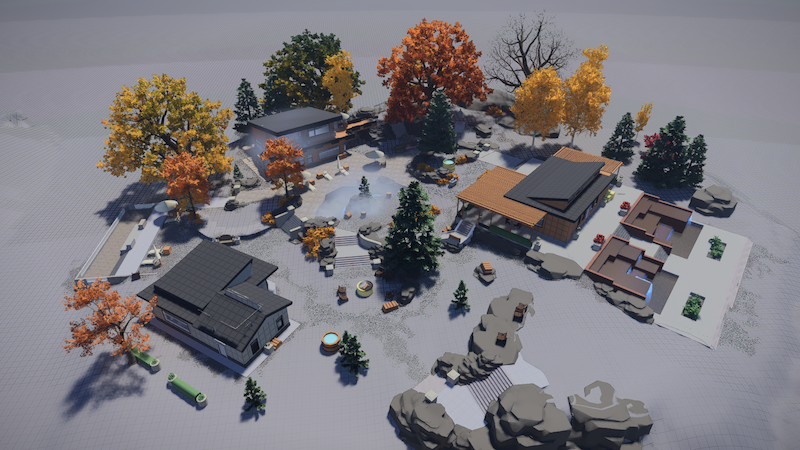}
    \caption{Environment A}
\end{subfigure}
\hfill
\begin{subfigure}{0.32\textwidth}
    \includegraphics[width=\textwidth]{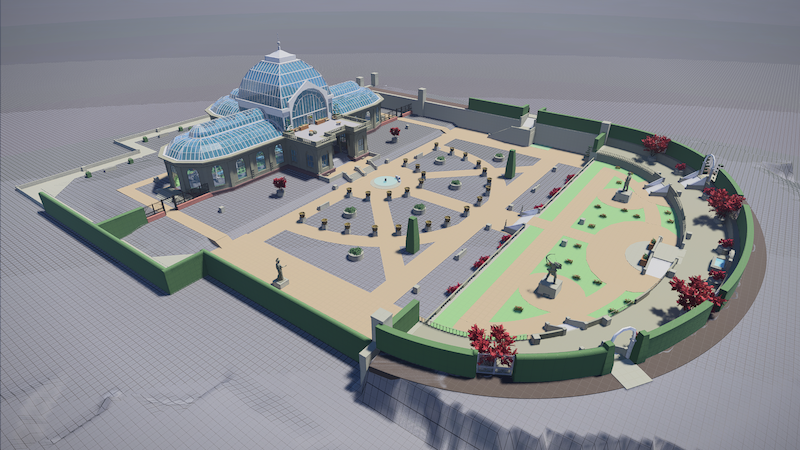}
    \caption{Environment B}
\end{subfigure}
\hfill
\begin{subfigure}{0.32\textwidth}
    \includegraphics[width=\textwidth]{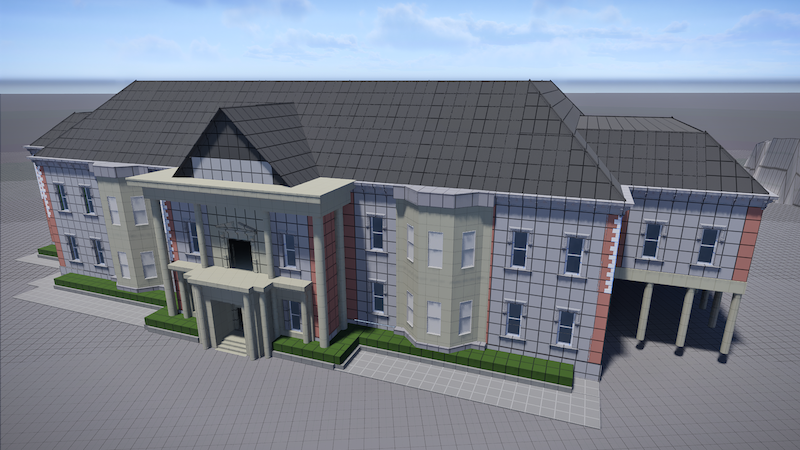}
    \caption{Environment C}
\end{subfigure}
        
\caption{Three types of scene for in-game evaluation. \textit{Environment A} represents a compact outdoor environment, \textit{Environment B} represents a wide outdoor environment, \textit{Environment C} represents a multi-level indoor environment.}
\label{fig:in_game_environment}
\end{figure}

\begin{table}
\footnotesize
  \caption{In-game evaluation results.}
  \label{tab:in-game_evaluaton}
  \begin{tabular}{l|ccc|ccc|ccc}
    \toprule
    \multirow{3}{*}{} &
      \multicolumn{3}{c|}{\textbf{Environment A} } &
      \multicolumn{3}{c|}{\textbf{Environment B}}&
      \multicolumn{3}{c}{\textbf{Environment C}} \\ &
      \multicolumn{3}{c|}{(809k triangles)} &
      \multicolumn{3}{c|}{(1078k triangles)}&
      \multicolumn{3}{c}{(136k triangles)} \\
      & {Cold (us)} & {Warm (us)} & {Acc.} & {Cold (us)} & {Warm (us)} & {Acc.} & {Cold (us)} & {Warm (us)} & {Acc.} \\
    \midrule
    Raycasting & 55.28 & 8.79 & - & 78.79 & 13.46 & - & 62.37 & 10.84  & - \\
    Ours \((128 \times 4)\) & 10.0 & 4.5 & 0.950 & 10.0 & 4.5 & 0.897 & 10.0 & 4.5  & 0.990\\
    Ours \((128 \times 2)\) & 7.0 & 2.3 & 0.950 & 7.0 & 2.3 & 0.894 & 7.0 & 2.3  & 0.989\\
    Ours \((64 \times 2)\) & 4.0 & 1.3 & 0.947 & 4.0 & 1.3 & 0.886 & 4.0 & 1.3  & 0.988\\
    Ours \((32 \times 2)\) & 3.0 & 1.1 & 0.940 & 3.0 & 1.1 & 0.881 & 3.0 & 1.1  & 0.985\\
    \bottomrule
  \end{tabular}
\end{table}

To evaluate our method in a real game environment, we perform a side-by-side comparison with raycasting using an AAA game engine. The raycasting implementation is based on a commonly used commercial product called \textit{Havok Physics}. The maximum raycasting distance for both data collection and baseline comparison is set to 100 meters. Three types of scenes are selected as our evaluation environment, as shown in Fig. \ref{fig:in_game_environment}. These scenes are partitioned into \SI{16}{\metre} \(\times\) \SI{16}{\metre} \(\times\) \SI{16}{\metre} voxels, and each partition contains a different number of source positions of interest, ranging from 30 to 200. The evaluation results for execution time and visibility prediction accuracy are presented in Table.\ref{tab:in-game_evaluaton}.  Due to the significant impact of CPU caching on execution time, we collect both cold start and warm start average execution times for all tasks. The execution time and accuracy of our method are measured based on a C++ implementation and executed on a single-threaded CPU process without batching. We use (PE, Long-Lat) version of our method. In addition, various sizes of MLPs are evaluated, while keeping the encoding settings the same (\(n_{\text{freq}}=6,\, F=2,\,L=16\)). So the input size of the MLPs is \(2 \times 3 \times 6 + 2 \times 16 = 68\). 

\begin{table}
\footnotesize
  \caption{In-game memory usage estimation.}
  \label{tab:in-game_memory}
  \begin{tabular}{l|c|c|cccc}
    \toprule
    & Raycasting & Part. \# & Ours \((128 \times 4)\) & Ours \((128 \times 2)\) & Ours \((64 \times 4)\) & Ours \((32 \times 2)\) \\
    \midrule
    \textbf{Environment A} & 13.3 MB & 56 & 104.7 MB & 97.3 MB & 95.4 MB & 92.3 MB \\
    \textbf{Environment B} & 30.5 MB & 80 & 149.5 MB & 139 MB & 136.2 MB & 131.9 MB \\
    \textbf{Environment C} & 6.9 MB & 30 & 56.1 MB & 52.1 MB & 51.1 MB & 49.5 MB \\
    \bottomrule
  \end{tabular}
\end{table}

For each partition, using 32-bit float precision, the memory size of the multi-resolution grid is approximately 1.64 MB, while the memory size of the MLP ranges from 13 KB to 234 KB. The observed memory usage is lower than that of offline experiments due to the utilization of smaller feature vectors and MLP for in-game evaluation. To compare memory usage with raycasting, as shown in Table \ref{tab:in-game_memory}, we estimate raycasting memory usage by assessing the memory size of the collision geometry at the level of detail employed for raycasting. For our methods, we estimate memory usage by multiplying the number of partitions for each environment by the model size for each partition.

Due to the different characteristics such as average collision distance and number of triangles in the three scenes, the execution time of raycasting-based visibility is different. In contrast, our inference time remains the same across the three scenes because we are using exactly the same model architecture. This is potentially a desirable property for game AI applications, as it makes the execution time more predictable.

\begin{figure}[hbt!]
\centering
\begin{subfigure}{0.537\textwidth}
    \includegraphics[width=\textwidth]{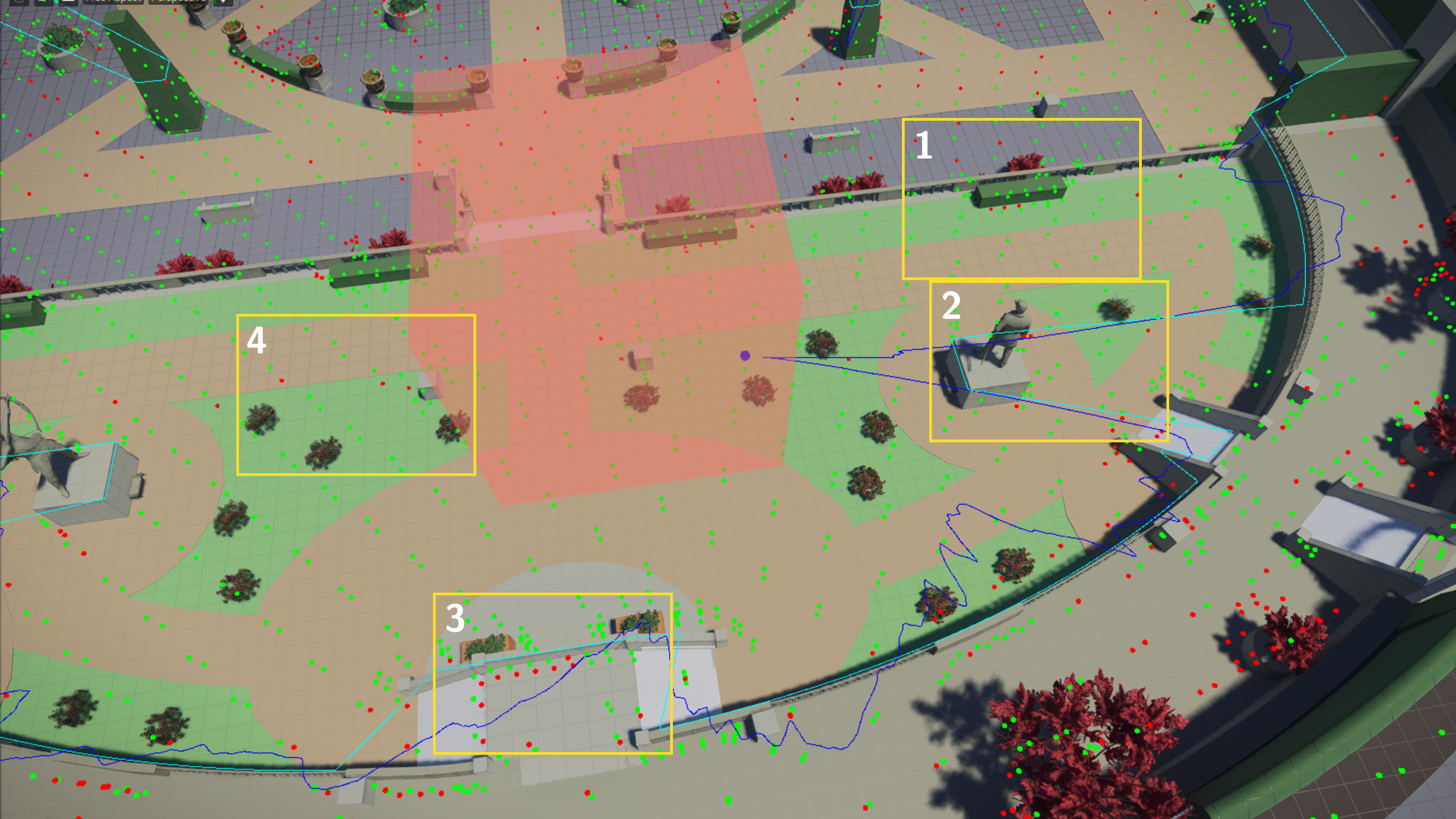}
    \caption{}
\end{subfigure}
\hfill
\begin{subfigure}{0.451\textwidth}
    \includegraphics[width=\textwidth]{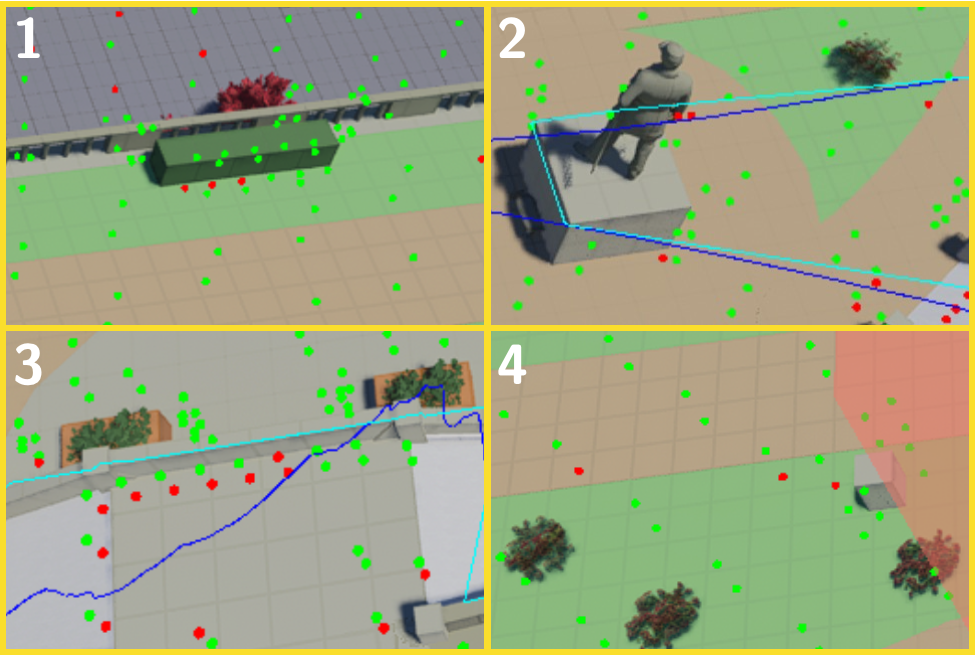}
    \caption{}
\end{subfigure}
        
\caption{Visualizations of visibility prediction in Environment B. \textbf{(a)} The selected partition is visualized as an orange voxel. The source position is visualized as a blue sphere inside this voxel. Visibility target positions that were predicted correctly or incorrectly are visualized as green and red spheres, respectively. \textbf{(b)} Highlighted areas of predictions.}
\label{fig:environment_2}
\end{figure}

From the experimental results presented in Table.\ref{tab:in-game_evaluaton}, we observe that the most challenging scene for our method is Environment B, which contains many small objects in the open area. We hypothesize that the reason for the lower accuracy is due to the reconstruction error at object edges. As observed in the visualizations shown in Fig. \ref{fig:environment_2}, most of the incorrect predictions are around object edges, corresponding to the viewing direction from the source position. In the future, this problem may solved by using a hybrid approach that detects object edges and switches between raycasting-based visibility tests and ours. For Environment C, where most of the target positions are not visible, the prediction accuracy of our method is very high (\(> 98.5\%\)). Under different MLP settings and scenes, our method shows speedups ranging from \(1.95 \times\) to \(26.26 \times\).

\section{Discussion and future work}
In this section, we will discuss the limitations of our method and possible directions for future work.

\textit{Direction feature enhancements}. In our method we use a multi-resolution grid for direction encoding and use it to store learnable, locally interpolated features. While our current techniques for sphere discretization and feature interpolation remain relatively unexplored, there exists potential for enhancement and refinement. To make the features more informative, in the future we can explore sphere discretization methods with better mathematical properties such as HEALPix \cite{gorski2005healpix}, data structures that take spherical curvature into account, such as Spherical Feature-Grid \cite{dou2023real} or adopt better feature interpolation methods that use the geodesic distance.

\textit{Precomputation time}. As a form of precomputation methods, our approach requires learning of the surface geometry information of the scene into another representation before being evaluated at runtime. A disadvantage of precomputation methods is the additional time required for offline processing. The precomputation time for our method consists of data collection time and model training time. For small areas of the game scene, this process takes several minutes, which is acceptable. However, scaling our method to the entire game world could extend the precomputation time to several hours or even days, depending on the number of positions of interest and directions. To improve the data collection process in the future, instead of relying on raycasting with predefined directions, we can explore the use of panoramic depth frames from the GPU. This improvement would allow us to efficiently collect omnidirectional distance data for positions, taking advantage of the graphics pipeline.

\textit{Dynamic scene handling} is another limitation of our method. This is because many game scenarios involve highly dynamic scenes, including destructible walls and buildings, large moving objects such as vehicles, dynamic visual effects such as smoke and fog, etc. However, our current method is only applicable to static part of scenes. To overcome this limitation, future improvements could include modeling dynamic objects with temporal variations \cite{ost2021scenegraphs, cao2023hexplane, luiten2023dynamic} or variations of the game scene.

\textit{Extended game applications to replace raycasting}. In addition to the visibility test use case presented in this paper, raycasting is widely used in many other aspects of video game development. For example, navigation mesh construction, path finding, projectile ballistics, spatial clearance tests, UI logic, and melee combat all rely heavily on raycasting techniques. While these applications may have different requirements in terms of memory, accuracy, and computational cost, by adjusting the grid and MLP settings, our method can be used as an accelerated approximation technique. It would be valuable to explore the potential in other application scenarios to replace traditional raycasting.

\textit{Generalized representation}. Spatial reasoning is a major challenge in crafting more sophisticated, believable, and intelligent NPC behavior in games. Existing approaches, such as \cite{forbus2002qualitative, dill2019spatial}, have made progress in this area. Our method provides a potential solution to their computational challenge by learning a neural representation of the spatial environment that is not limited to ray and distance pairs. We can generalize the representation to incorporate other information obtained from game environment for efficient information retrieval at runtime. We see many potential applications of our approach in this area.

\section{Conclusion}
For game AI applications, obtaining real-time visibility information is critical but computationally expensive. Our work presents the first method designed to address the computational bottleneck of visibility tests for game AI using a neural representation.

We show that by optimizing neural networks to represent scene geometry as an ODF, we can efficiently predict the visibility between positions. In addition, evaluating visibility using our representation only depends on the local ODF associated with the source position, regardless of the location of the target. This allows scalability to the entire game world without sacrificing model inference time, at the cost of potentially larger overall memory requirements. We show that by using multi-resolution grid encoding for direction features, we can improve the expressiveness of our neural network for ODF reconstruction. Compared to raycasting in a real game environment, we not only achieve speed up, but also provide constant time inference. And besides, as an efficient representation designed for game AI applications, our method can potentially serve as a fast information retrieval solution for complex systems such as spatial reasoning in the future.

\begin{acks}
We would like to thank Georges Nader, Ludovic Denoyer, Jean-Philippe Barrette-LaPierre, Alexis Rolland, and Yves Jacquier for their valuable discussions and reviews. This work was supported by Ubisoft.
\end{acks}

\bibliographystyle{ACM-Reference-Format}
\bibliography{ref}

\appendix

\end{document}